\documentclass[11pt]{article}

\usepackage[preprint]{acl}
\usepackage{tabularx}
\usepackage{times}
\usepackage{latexsym}
\usepackage{booktabs}
\usepackage{subcaption}
\usepackage{float}
\usepackage[T1]{fontenc}

\usepackage[utf8]{inputenc}

\usepackage{microtype}

\usepackage{inconsolata}

\usepackage{graphicx}

\title{TopicENA: Enabling Epistemic Network Analysis at Scale through Automated Topic-Based Coding}

\author{Owen H.T. Lu \\
  National Chengchi University  \\
  \texttt{owen.lu.academic@gmail.com} \\\And
  Tiffany T.Y. Hsu \\
  The London School of Economics \\and Political Science \\
  \texttt{tyhsu.ac@gmail.com} \\}

\begin{document}

\maketitle
\begin{abstract}

Epistemic Network Analysis (ENA) is a method for investigating the relational structure of concepts in text by representing co-occurring concepts as networks. Traditional ENA, however, relies heavily on manual expert coding, which limits its scalability and real-world applicability to large text corpora. Topic modeling provides an automated approach to extracting concept-level representations from text and can serve as an alternative to manual coding. To tackle this limitation, the present study merges BERTopic with ENA and introduces TopicENA, a topic-based epistemic network analysis framework. TopicENA substitutes manual concept coding with automatically generated topics while maintaining ENA’s capacity for modeling structural associations among concepts. To explain the impact of modeling choices on TopicENA outcomes, three analysis cases are presented. The first case assesses the effect of topic granularity, indicating that coarse-grained topics are preferable for large datasets, whereas fine-grained topics are more effective for smaller datasets. The second case examines topic inclusion thresholds and finds that threshold values should be adjusted according to topic quality indicators to balance network consistency and interpretability. The third case tests TopicENA’s scalability by applying it to a substantially larger dataset than those used in previous ENA studies. Collectively, these cases illustrate that TopicENA facilitates practical and interpretable ENA analysis at scale and offers concrete guidance for configuring topic-based ENA pipelines in large-scale text analysis.

\end{abstract}

\section{Introduction}

Epistemic Network Analysis (ENA) is a learning analytics method designed to examine how ideas, actions, or cognitive elements are connected within learning and interaction data. Instead of focusing on how frequent concepts appear, ENA emphasizes how different elements co-occur within meaningful contexts, and displaying these relationships as networks. By transforming discourse feature into network structures, ENA allows researchers to analyze and statistically compare how learners or groups organize their knowledge and problem solving processes~\cite{shaffer2016tutorial}. Recent studies have applied ENA to a wide range of educational contexts to investigate what learners actually integrate during learning activities. For example, \citet{tu2025effects} used ENA to examine how students with different levels of digital self efficacy developed distinct conceptual connections across multiple stages of reflection in a data literacy course, helping to explain differences in learning achievement and learning approaches. \citet{chang2025generative} applied ENA to reflective writing supported by generative AI and showed how higher order thinking concepts were connected differently when students engaged with AI assisted reflection. Other studies have used ENA to analyze Self-Regulated Learning (SRL) behaviors in open-ended problem solving environments~\cite{paquette2021using}, and to examine teachers beliefs and professional practices in authentic educational settings~\cite{pantic2022making}. Previous studies demonstrate ENA as a powerful approach for revealing underlying cognitive and epistemic structures that are not visible through traditional outcome based analyses.

Previous studies using ENA largely rely on expert-based manual or semi-manual coding of textual data, which constrains the scale of analysis. For example, \citet{ko2024exploring} treated each forum post as a unit of analysis and coded a total of 390 discussion posts for ENA modeling. In more qualitatively oriented studies, the number of coded units is often smaller. In a study on pre-service teachers’ identity development, researchers conducted multiple rounds of expert coding on interview data and analyzed only 143 utterances using ENA \cite{vega2021negotiating}. Similarly, \citet{chang2025generative} applied ENA to students’ reflective writing and manually coded 44 reflection reports across experimental and control groups. Even in recent large-scale classroom studies, manual coding remains a key bottleneck. For instance, \citet{tu2025effects} coded students’ reflective reports collected at three time points (beginning, midterm, and end of the course), requiring experienced coders to reach acceptable inter-rater agreement. Taken together, these studies indicate that while ENA provides fine-grained insights into learners’ knowledge structures, its strong dependence on expert coding limits its scalability for large text datasets.

In recent years, some researchers have tried to lessen the reliance of ENA on manual expert coding by introducing automated coding methods to support large-scale analysis. For example, \citet{cai2017epistemic} applied topic modeling to automatically extract codes from chat data in a joint learning environment and analyzed more than 15,000 utterances, which was several times larger than the scale of most manually coded ENA studies. Similarly, \citet{gavsevic2019sens} combined Latent Dirichlet Allocation (LDA) with ENA to conduct automated content analysis of more than 6,000 learner-generated posts, using topics as coding units to represent cognitive dimensions in epistemic networks. These studies show that topic modeling-based automatic coding can substantially increase the scale of data that ENA can handle and extend its applicability to large text corpora. However, existing studies are often designed for specific research contexts and provide limited methodological guidance on how automated topic-based coding choices influence ENA results, especially under different data scales. To address this gap, the present study introduces TopicENA, a generalizable analysis tool that integrates topic modeling with ENA to support automated semantic coding and scalable discourse analysis. Beyond proposing a tool, this study thoroughly examines how key design choices in topic modeling affect ENA outcomes across three analytic cases. Based on these cases, the study shows how TopicENA can be configured and interpreted across different data contexts, thereby supporting both large-scale and at-scale ENA analysis. This paper makes the following contributions:

\begin{itemize}
    \item This study proposes TopicENA, a generalizable framework that integrates topic modeling with ENA to enable automated semantic coding and scalable discourse analysis without relying on manual expert coding.

    \item Through three analytic cases, this study systematically examines how topic granularity, topic inclusion thresholds, and data scale influence ENA results, providing methodological guidance for configuring topic-based ENA analyses.
    
    \item By shifting expert involvement from instance-level coding to higher-level interpretation and analytic decision making, TopicENA supports more interpretable and reproducible ENA applications in large text corpora.

\end{itemize}

\section{Related Work} \label{sec:related_works}

Topic modeling is an unsupervised text analysis approach that aims to automatically extract latent semantic topic structures from large collections of documents. The assumption of this approach is that each document is composed of multiple latent topics, and that each topic is represented by a group of words with similar semantic meanings. By using topic modeling, researchers can summarize and explore the semantic patterns of large-scale textual data without manual coding. Therefore, topic modeling has been widely applied across fields such as social sciences, educational research, information science, and digital media analysis~\cite{blei2012probabilistic, grimmer2013text}.

Among various topic modeling approaches, LDA has been regarded as the most representative classical model. For example, \citet{gavsevic2019sens} employed LDA to generate topic-based codes and used them as inputs to ENA to visualize the structure of learners’ discourse. LDA is built on a hierarchical Bayesian probabilistic framework, in which each document is modeled as a mixture of multiple topics, and every topic is represented by a probability distribution over words~\cite{blei2003latent}. Due to its clear mathematical assumptions and stable model structure, LDA has been the dominant topic modeling method for more than a decade and has been applied extensively in long text analysis and theory-driven social science research~\cite{kherwa2020topic}. However, recent reviews have highlighted several limitations of LDA. Because LDA relies on the bag-of-words assumption, it cannot effectively preserve contextual information, which leads to reduced topic quality and interpretability when analyzing short texts or texts with similar meanings but different word usage. In addition, LDA’s performance is highly dependent on the number of topics, and its outputs often require substantial manual interpretation and post-processing, increasing the subjectivity and workload for researchers~\cite{chen2023we, kherwa2020topic}.

To address the limitations of LDA, recent studies have shifted toward neural topic modeling approaches that integrate word embeddings and deep learning, with BERTopic gaining growing attention. BERTopic uses pre-trained language models such as Sentence-BERT(SBERT) to generate semantic embeddings, followed by dimensionality reduction with UMAP and density-based clustering with HDBSCAN to identify topics. Topic representations are then summarized using a class-based TF-IDF procedure~\cite{grootendorst2022bertopic}. Compared with traditional bag-of-words-based methods, BERTopic better preserves contextual information and can be applied to both long and short texts, including social media posts, online comments, and other forms of user-generated content, while producing coherent topic representations. Recent studies have shown that BERTopic performs consistently across different text lengths and scales well to large datasets, making it suitable for large-scale text analysis~\cite{janssens2025comparative}. Moreover, BERTopic has been increasingly adopted in systematic literature reviews and cross-domain topic analysis, as demonstrated in the analysis of AI research trends reported in~\citet{raman2024unveiling}. Overall, embedding-based topic modeling, particularly BERTopic, has become an important approach for analyzing large scale and unstructured textual data~\cite{chen2023we, grimmer2013text}.

To sum up, the focus of topic modeling research has gradually shifted from simple semantic summarization to approaches that emphasize semantic quality and support for short texts, while leveraging neural networks to achieve improved topic detection. Therefore, this study adopts BERTopic as the automated semantic coding tool in the TopicENA framework. By using its stable and context-aware topic representations, textual data can be transformed into conceptual units that support subsequent structural relationship modeling with ENA, thereby enabling large-scale analysis tasks.

\section{TopicENA} \label{sec:topic_ena}

\subsection{Overview}

The proposed TopicENA is an automated discourse analysis framework that integrates neural topic modeling with ENA to support scalable modeling of conceptual relationships in large textual corpora. Instead of relying on theory-driven or manually designed coding schemes, TopicENA employs probabilistic topic representations as semantic units for ENA, enabling large-scale discourse data to be systematically transformed into epistemic networks. The framework is designed as a sequential analysis pipeline consisting of three main stages.
\begin{itemize}

    \item \textbf{Neural Topic Induction and Multi-Topic Attribution}:
    In the initial stage, TopicENA applies BERTopic to extract latent semantic topics from the input corpus by employing semantic embedding, dimensionality reduction, and clustering. Each document is characterized as a probabilistic distribution across multiple topics, rather than being restricted to a single category. This approach preserves semantic overlap among documents and reflects the complex, multi-dimensional nature of discourse. The degree of semantic abstraction for topic induction is determined by the topic granularity specified in the topic modeling configuration, which affects the number, size, and specificity of the generated topics. A detailed discussion of topic granularity appears in Section~\ref{sec:topic_granularity}.
    
    \item \textbf{Topic-to-ENA Encoding}:  
    In the second stage, the probabilistic topic representations are converted into ENA-compatible encoding units. Topics are served as candidate semantic elements, and their presence within each document is determined based on probabilistic criteria. This encoding process transforms soft topic representations into structured co-occurrence data that can be used for model conceptual relations, addressing a key limitation of conventional ENA pipelines in which one-to-one document coding restricts the formation of meaningful networks. This inclusion decision is controlled by a topic inclusion threshold, which determines whether a topic is considered present in a document for ENA encoding; further details are described in Section~\ref{sec:topic_inclusion_threshold}.
    
    \item \textbf{Epistemic Network Construction and Visualization}:  
    In the final stage, the encoded topic data are processed using an R-based ENA pipeline to construct networks that represent the relational structure among topics. TopicENA supports automated generation of network models across different analytical conditions or comparison groups, producing visualizations for the examination and comparison of knowledge structures at-scale.

\end{itemize}

\subsection{Topic granularity}\label{sec:topic_granularity}

In TopicENA, topic granularity refers to the level of semantic abstraction at which latent topics are induced from the textual corpus. Topic granularity is primarily controlled through the configuration of the topic induction process in BERTopic, which relies on dimensionality reduction and density-based clustering to identify coherent semantic groups.

Specifically, topic granularity is affected by parameters from both UMAP and HDBSCAN. Within HDBSCAN, the parameters \texttt{min\_cluster\_size} and \texttt{min\_samples} jointly regulate the size and stability of topic clusters. Reducing \texttt{min\_cluster\_size} allows smaller and more fine-grained topics to emerge, while lowering \texttt{min\_samples} relaxes cluster density requirements, making topic formation more permissive but potentially increasing noise.

UMAP parameters influence topic granularity by determining the preservation of semantic structure during dimensionality reduction. The parameter \texttt{n\_neighbors} regulates the balance between local and global structure; smaller values highlight local semantic distinctions and facilitate the separation of fine-grained topics. The parameter \texttt{n\_components} defines the dimensionality of the reduced embedding space, with higher values retaining more semantic information for subsequent clustering. Additionally, \texttt{min\_dist} affects the compactness of document groupings in the embedding space, where smaller values promote denser clusters of semantically similar texts.

The \texttt{min\_topic\_size} parameter in BERTopic sets the smallest number of documents needed for a topic to be kept. It works as a filter, removing topics that are too small or unstable, which affects how detailed the final set of topics will be.

\subsection{Topic inclusion threshold}\label{sec:topic_inclusion_threshold}

Topic granularity affects how topics are identified, but TopicENA also needs a way to encode these topics for epistemic network analysis. Since a single document can cover multiple themes, TopicENA uses a probabilistic topic-inclusion strategy rather than assigning a single topic to each document.

BERTopic produces a probability distribution that represents the association strength between each document and each induced topic. In typical BERTopic usage, each document is assigned to a single dominant topic, resulting in a one-hot topic representation that reflects only the most probable topic per document. TopicENA introduces a topic inclusion threshold, implemented as the parameter \texttt{topic\_inclusion\_th}, to determine whether a topic is considered present in a document for ENA encoding. A topic is included if its probability exceeds the specified threshold.

With this threshold-based approach, a document can be linked to multiple topics, allowing them to co-occur within the same document. This co-occurrence is important for building useful epistemic networks, as ENA examines how semantic elements co-occur. The topic inclusion threshold turns probabilistic topic data into clear inclusion decisions, connecting neural topic modeling results to ENA-ready co-occurrence data.

\section{Experimental Setup} \label{sec:experimental_setup}

\subsection{Dataset}

The ASAP (Automated Student Assessment Prize) 2.0 dataset \cite{learning-agency-lab-automated-essay-scoring-2} is a large open-access corpus comprising seven assignments and 24,728 source-based argumentative essays written by U.S. secondary school students. All essays were holistically scored by trained human raters using a unified 1–6 scale. Reported inter-rater reliability prior to adjudication ranged from weighted Cohen’s $\kappa$ = 0.659 to 0.745, indicating acceptable to good agreement for large-scale writing assessment \cite{crossley2025large}.

\subsection{Experimental Procedure} \label{sec:experimental_procedure}

\subsubsection{Dataset Selection and Preparation}

The ASAP 2.0 dataset comprises seven assignments, each with a different number of student essays stored in the \texttt{full\_text} column. For utterance-level analysis, essays are segmented into sentence-level units using periods as boundary markers, with each sentence treated as a minimal discourse unit for co-occurrence analysis. Through this preprocessing step, a total of 457,002 utterances were obtained across the dataset. The corresponding numbers of essays and utterances for each assignment are summarized in Table~\ref{tab:asap_doc_number}.

\begin{table}[ht]
    \setlength{\tabcolsep}{4pt} 
    \begin{tabularx}{\columnwidth}{ccc}
    \toprule
    Assignment No.     & Essay & Utterance  \\
    \midrule
    1       & 4,480 & 72,463         \\
    2       & 4,883 & 88,282         \\
    3       & 6,170 & 53,226         \\
    4       & 2,046 & 37,173         \\
    5       & 1,959 & 119,052        \\
    6       & 2,175 & 40,925         \\
    7       & 3,015 & 45,881         \\
    \midrule
            & 24,728    & 457,002    \\
    \bottomrule
    \end{tabularx}
    \caption{Essay and utterance count in the ASAP 2.0 dataset.}
    \label{tab:asap_doc_number}
\end{table}

To align with the experimental design of this study, the ASAP 2.0 dataset was not used in a single pass. Instead, progressively selected subsets of the dataset were used to examine the sensitivity and scalability of TopicENA under varying analytical conditions. Specifically, the three experimental cases focus on (1) sensitivity to topic granularity, (2) sensitivity to topic inclusion threshold, and (3) scalability in large-scale discourse analysis. Accordingly, assignments with different data scales were used at different stages of the experiments.

We first started with a smaller-scale assignment to examine the sensitivity of TopicENA to parameter settings under relatively constrained data conditions. \textit{\textbf{Assignment 4}} was selected for this purpose, as it contains the fewest utterances in the ASAP 2.0 dataset. \textit{\textbf{Assignment 4}} includes 2,046 essays, which were segmented into a total of 37,173 utterances after sentence-level preprocessing. The assignment prompt is as follows:

\begin{quote}
    \textit{\textbf{Assignment 4}: Write a letter to your state senator in which you argue in favor of keeping the Electoral College or changing to election by popular vote for the president of the United States. Use the information from the texts in your essay. Manage your time carefully so that you can read the passages; plan your response; write your response; and revise and edit your response. Be sure to include a claim; address counterclaims; use evidence from multiple sources; and avoid overly relying on one source. Your response should be in the form of a multiparagraph essay. Write your response in the space provided.}
\end{quote}

Following the analysis of \textit{\textbf{Assignment 4}}, we further evaluated TopicENA using a moderately sized dataset to examine whether the observed parameter sensitivities remained stable as the data scale increased. In this case, we selected \textit{\textbf{Assignment 5}}. \textit{\textbf{Assignment 5}} contains 1,959 essays, which were segmented into a total of 119,052 utterances. The assignment prompt is as follows:

\begin{quote}
    \textit{\textbf{Assignment 5} Write an explanatory essay to inform fellow citizens about the advantages of limiting car usage. Your essay must be based on ideas and information that can be found in the passage set. Manage your time carefully so that you can read the passages; plan your response; write your response; and revise and edit your response. Be sure to use evidence from multiple sources; and avoid overly relying on one source. Your response should be in the form of a multiparagraph essay. Write your essay in the space provided.}
\end{quote}

In addition, the ASAP 2.0 dataset provides expert assigned holistic scores. For the purposes of analysis, essays receiving scores from 1 to 3 were categorized as \texttt{LOW}, while essays receiving scores from 4 to 6 were categorized as \texttt{HIGH}. These two groups were used as contrasting groups for subsequent comparative assessments.

\subsubsection{Experimental Cases and Evaluation Strategy}

This study set up three experiments to examine how TopicENA behaves under different conditions. Each experiment focuses on a specific part of the TopicENA process.

In the first case (hereafter, \textbf{Case 1}), the experiment focuses on TopicENA's sensitivity to topic granularity. Topic granularity is controlled by adjusting parameters for both the UMAP and HDBSCAN components during the topic modelling stage. Specifically, the HDBSCAN parameters \texttt{min\_cluster\_size} and \texttt{min\_samples} are varied to regulate the size and stability of topic clusters, thereby influencing whether broader or more fine-grained topics are produced. In addition, UMAP parameters including \texttt{n\_neighbors}, \texttt{n\_components}, and \texttt{min\_dist} are adjusted to modify how local and global semantic structures are preserved in the reduced embedding space. By systematically varying these parameters, \textbf{Case 1} examines how changes in topic granularity are reflected in the resulting epistemic networks generated by ENA.

In the second case (hereafter, \textbf{Case 2}), the experiment investigates TopicENA's sensitivity to the topic inclusion threshold. This case focuses on the parameter \texttt{topic\_inclusion\_th}, which determines whether a topic is considered present in a document for ENA encoding based on its topic probability. By applying different threshold values, this case examines how stricter or more permissive topic inclusion criteria affect the construction, stability, and interpretability of epistemic networks. Similar to \textbf{Case 1}, this analysis is conducted across multiple assignments to observe how topic inclusion threshold settings interact with different discourse contexts and to derive practical guidelines for parameter tuning.

In the third case (hereafter: \textbf{Case 3}), the analysis of the entire ASAP 2.0 dataset, which includes all seven assignments. The main goal is to test if TopicENA can handle large-scale discourse analysis. This case tests whether TopicENA can find a clear set of 7 topics that match the assignments and build valid epistemic networks, including subtract networks, at scale. It serves as a stress test to determine whether TopicENA is practical and reliable for real-world, large-scale use.

\section{Results}

\subsection{Case 1: Sensitivity to Topic Granularity}

\textbf{Case 1} examines how topic granularity influences the results of TopicENA. Since TopicENA relies on BERTopic for topic modeling, we adjusted the \texttt{n\_neighbors}  parameter to control topic granularity. A larger \texttt{n\_neighbors}  produces coarser topics with fewer topic clusters, where each topic contains more documents. In contrast, a smaller \texttt{n\_neighbors}  produces finer topics with more clusters, where each topic contains fewer documents. In this study, the coarse, medium, and fine granularity settings correspond to \texttt{n\_neighbors} values of 60, 35, and 18, respectively. This case compares the effects of different granularity settings on ENA results using \textit{\textbf{Assignment 4}} and \textit{\textbf{Assignment 5}}.

\begin{figure*}[t]
    \centering

    \begin{subfigure}[t]{0.32\linewidth}
        \centering
        \includegraphics[width=\linewidth]{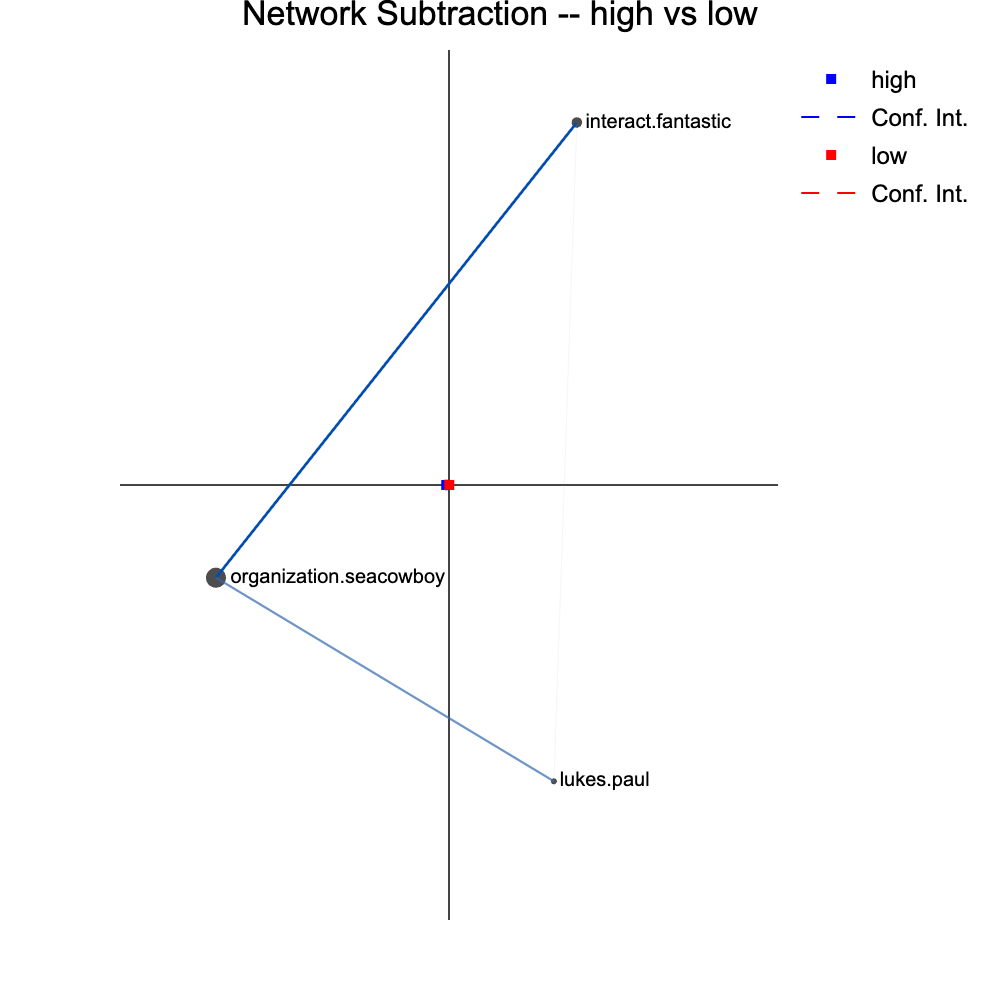}
        \caption{\textit{\textbf{Assignment 4}}, Coarse}
        \label{fig:case1_a4_coarse}
    \end{subfigure}\hfill
    \begin{subfigure}[t]{0.32\linewidth}
        \centering
        \includegraphics[width=\linewidth]{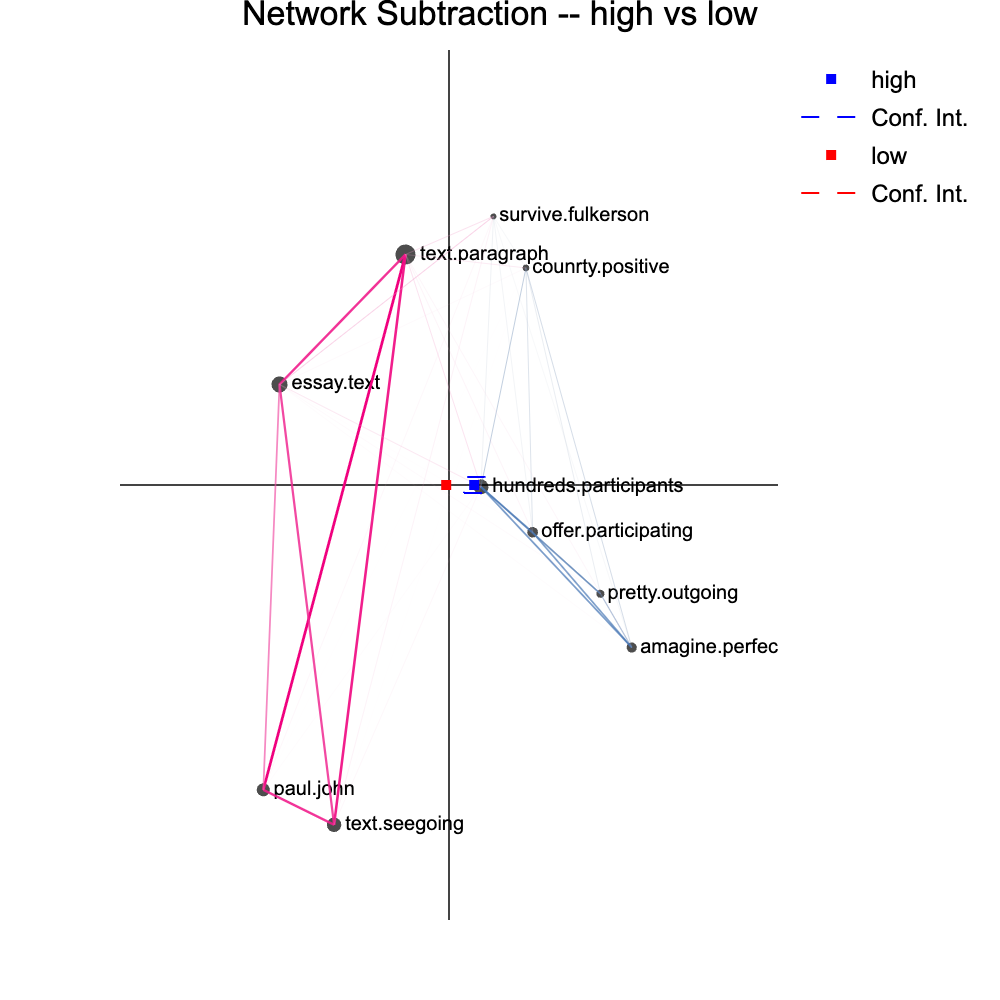}
        \caption{\textit{\textbf{Assignment 4}}, Medium}
        \label{fig:case1_a4_medium}
    \end{subfigure}\hfill
    \begin{subfigure}[t]{0.32\linewidth}
        \centering
        \includegraphics[width=\linewidth]{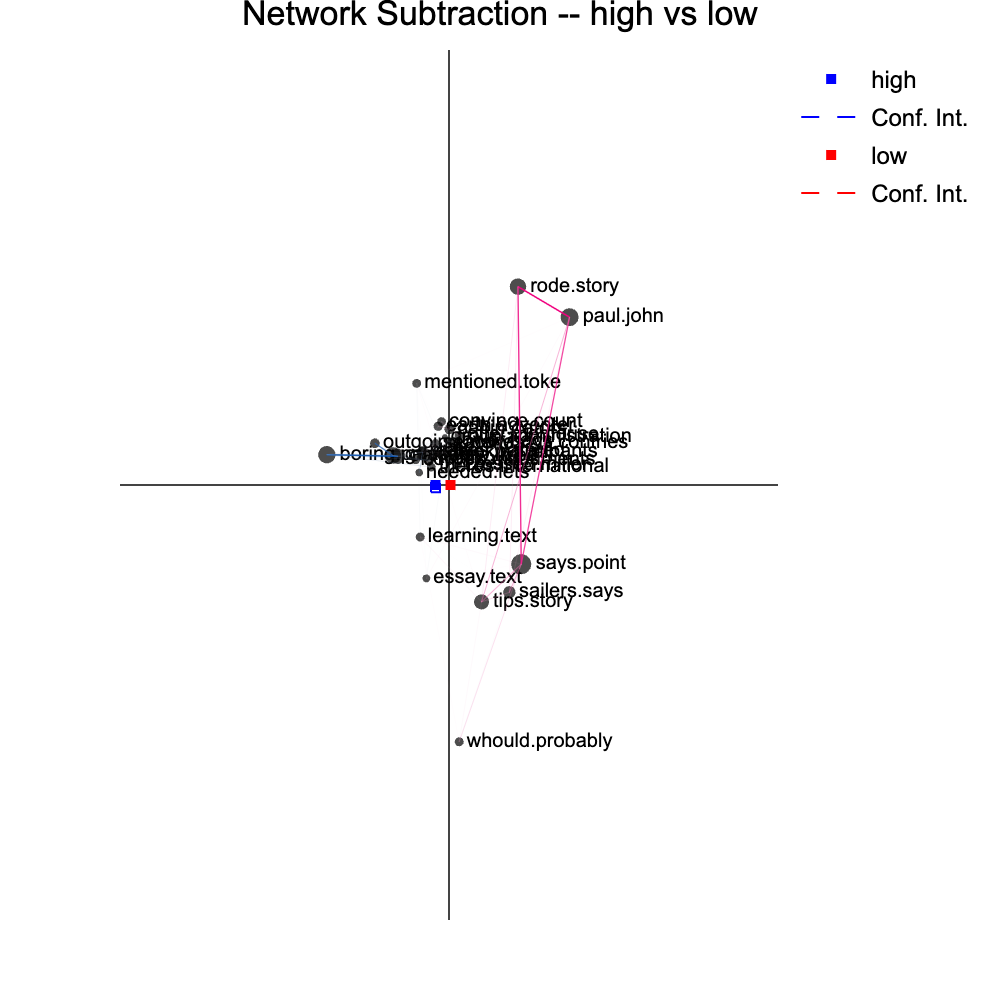}
        \caption{\textit{\textbf{Assignment 4}}, Fine}
        \label{fig:case1_a4_fine}
    \end{subfigure}

    \medskip

    \begin{subfigure}[t]{0.32\linewidth}
        \centering
        \includegraphics[width=\linewidth]{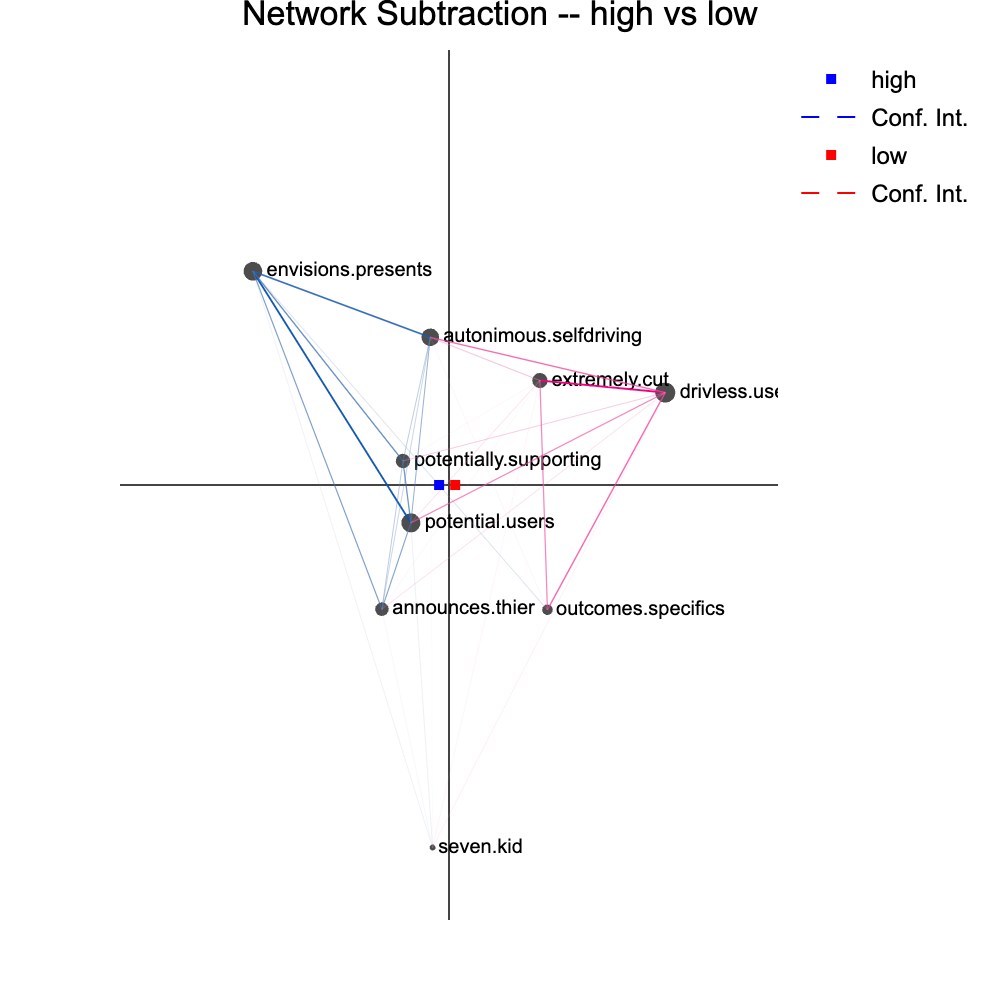}
        \caption{\textit{\textbf{Assignment 5}}, Coarse}
        \label{fig:case1_a5_coarse}
    \end{subfigure}\hfill
    \begin{subfigure}[t]{0.32\linewidth}
        \centering
        \includegraphics[width=\linewidth]{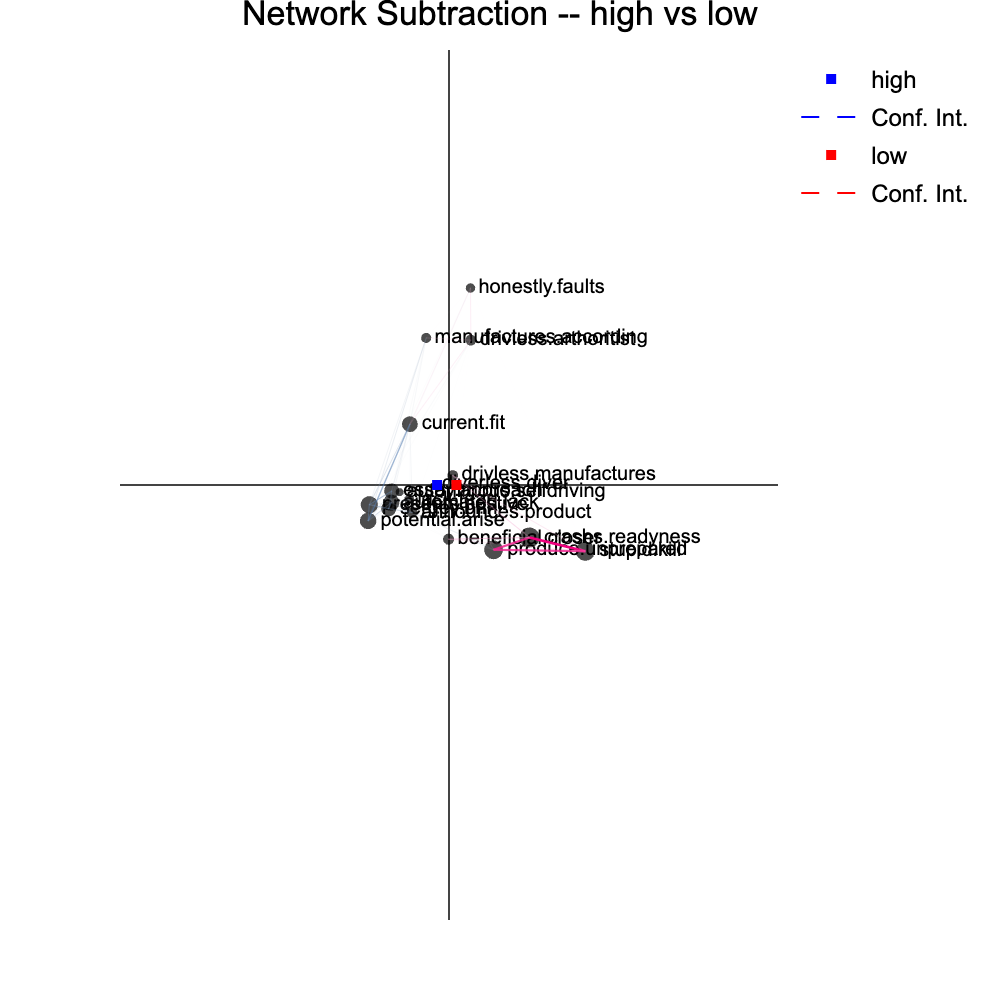}
        \caption{\textit{\textbf{Assignment 5}}, Medium}
        \label{fig:case1_a5_medium}
    \end{subfigure}\hfill
    \begin{subfigure}[t]{0.32\linewidth}
        \centering
        \includegraphics[width=\linewidth]{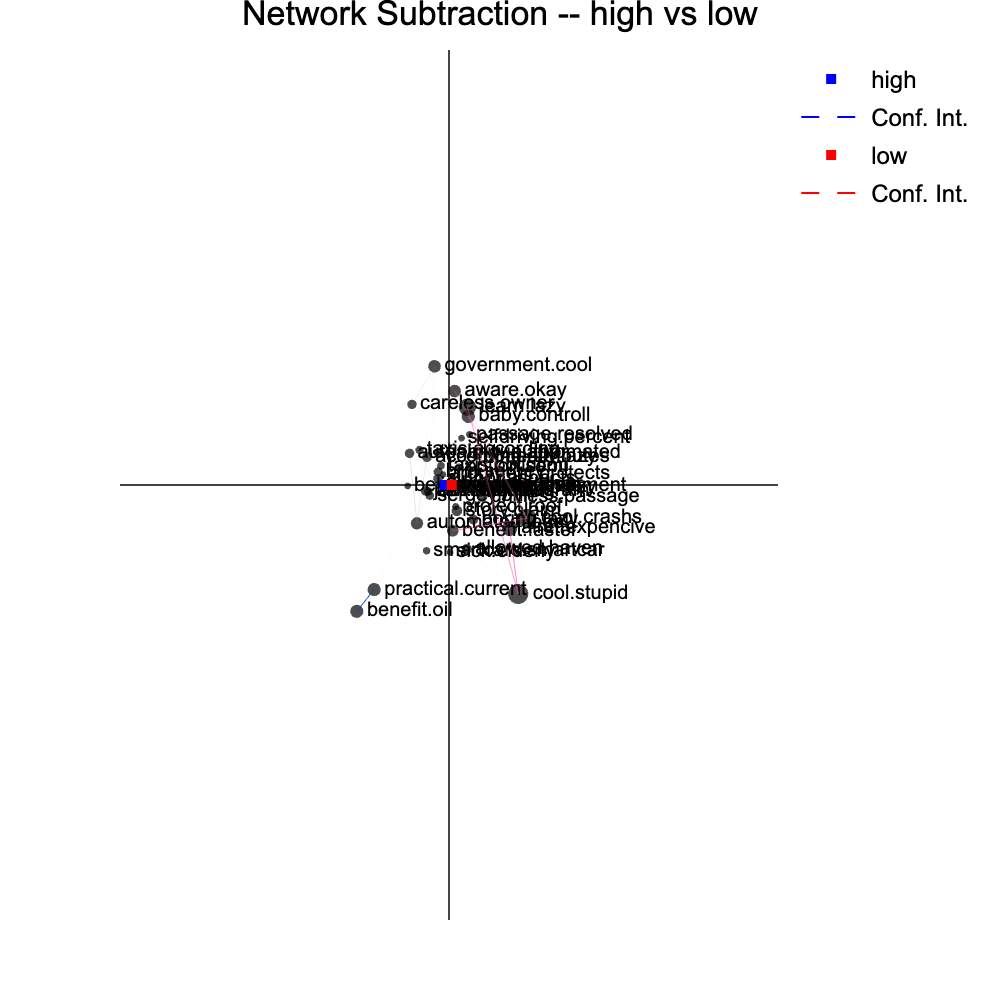}
        \caption{\textit{\textbf{Assignment 5}}, Fine}
        \label{fig:case1_a5_fine}
    \end{subfigure}

    \caption{ENA results under different topic granularity settings (coarse, medium, and fine) in \textbf{Case 1}}
    \label{fig:case1_result_2x3}
\end{figure*}

As shown in Figures~\ref{fig:case1_a4_coarse}, \ref{fig:case1_a4_medium}, and \ref{fig:case1_a4_fine}, \textit{\textbf{Assignment 4}}, which contains fewer utterances, is analyzed using coarse, medium, and fine topic granularity settings, respectively. As a result, the number of semantic nodes available for ENA was limited. Under this condition, the subtract network of the high-score group and the low-score group largely overlapped, making group differences difficult to interpret. When the granularity was adjusted to a medium level, the balance between the number of topics and the distribution of documents improved. Clearer structural differences between the high-score and low-score groups began to emerge in the ENA difference networks. However, when the granularity was further refined, the number of topics increased substantially while the number of documents per topic decreased. This led to sparse co-occurrence relations, and the resulting ENA networks became dense and overlapped, reducing their interpretability.

As shown in Figures~\ref{fig:case1_a5_coarse}, \ref{fig:case1_a5_medium}, and \ref{fig:case1_a5_fine}, \textit{\textbf{Assignment 5}}, which contains a larger volume of utterances, is presented under coarse, medium, and fine granularity settings, respectively. As the results, clearer ENA networks were observed even under coarse granularity settings. The larger data size ensured that each topic still included enough documents to support stable co-occurrence patterns. As a result, structural differences between the high-score and low-score groups remained visible. When medium granularity was applied, the increased number of topics further enhanced group-level structural differences, yielding the most interpretable ENA results. Similar to \textit{\textbf{Assignment 4}}, overly fine granularity produced too many topics with insufficient document support, which weakened the clarity of the ENA networks.

In summary, \textbf{Case 1} demonstrates that topic granularity should be aligned with data scale. For large datasets, coarser granularity helps maintain sufficient co-occurrence strength and improves the stability and readability of ENA results. For smaller datasets, moderately finer granularity can preserve semantic distinctions, but excessive refinement may reduce interpretability. These findings support the principle that large datasets benefit from coarse granularity, while smaller datasets benefit from finer granularity.

    


\subsection{Case 2: Sensitivity to Topic Inclusion Threshold}

\textbf{Case 2} investigates how the topic inclusion threshold, denoted as \texttt{topic\_inclusion\_th}, affects TopicENA results. After topic modeling, BERTopic assigns each document a topic score that represents the strength of association between the document and a topic. If a strict one-to-one topic assignment is used, each document is assigned to only one topic, which does not reflect the fact that a single text often addresses multiple concepts. To address this issue, we applied different probability thresholds to allow documents to be assigned to multiple topics when their topic scores exceeded the specified threshold. In \textbf{Case 2}, \texttt{topic\_inclusion\_th} was systematically varied across three levels, namely low (0.01), medium (0.05), and high (0.10), to examine how different inclusion strictness levels influence TopicENA results.

\begin{figure*}[t]
    \centering

    \begin{subfigure}[t]{0.32\linewidth}
        \centering
        \includegraphics[width=\linewidth]{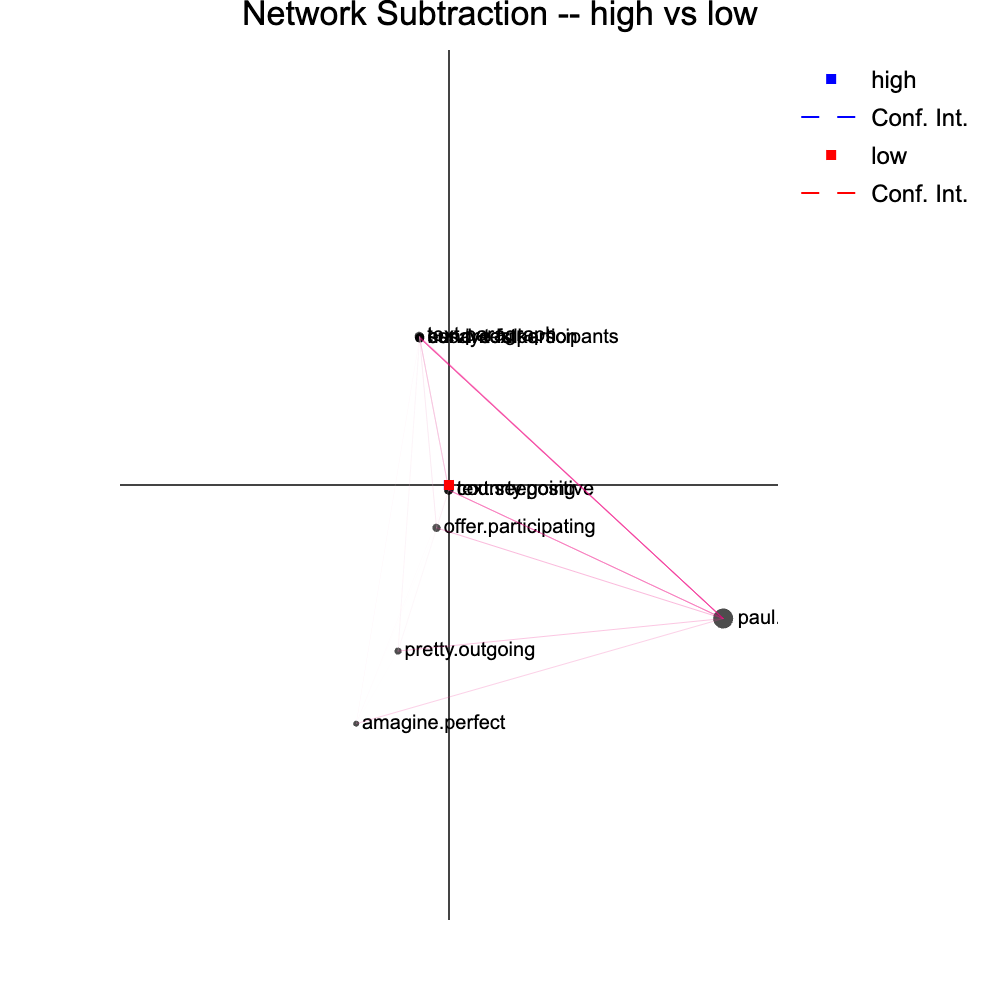}
        \caption{\textit{\textbf{Assignment 4}}}
        \label{fig:case2_a4_low}
    \end{subfigure}\hfill
    \begin{subfigure}[t]{0.32\linewidth}
        \centering
        \includegraphics[width=\linewidth]{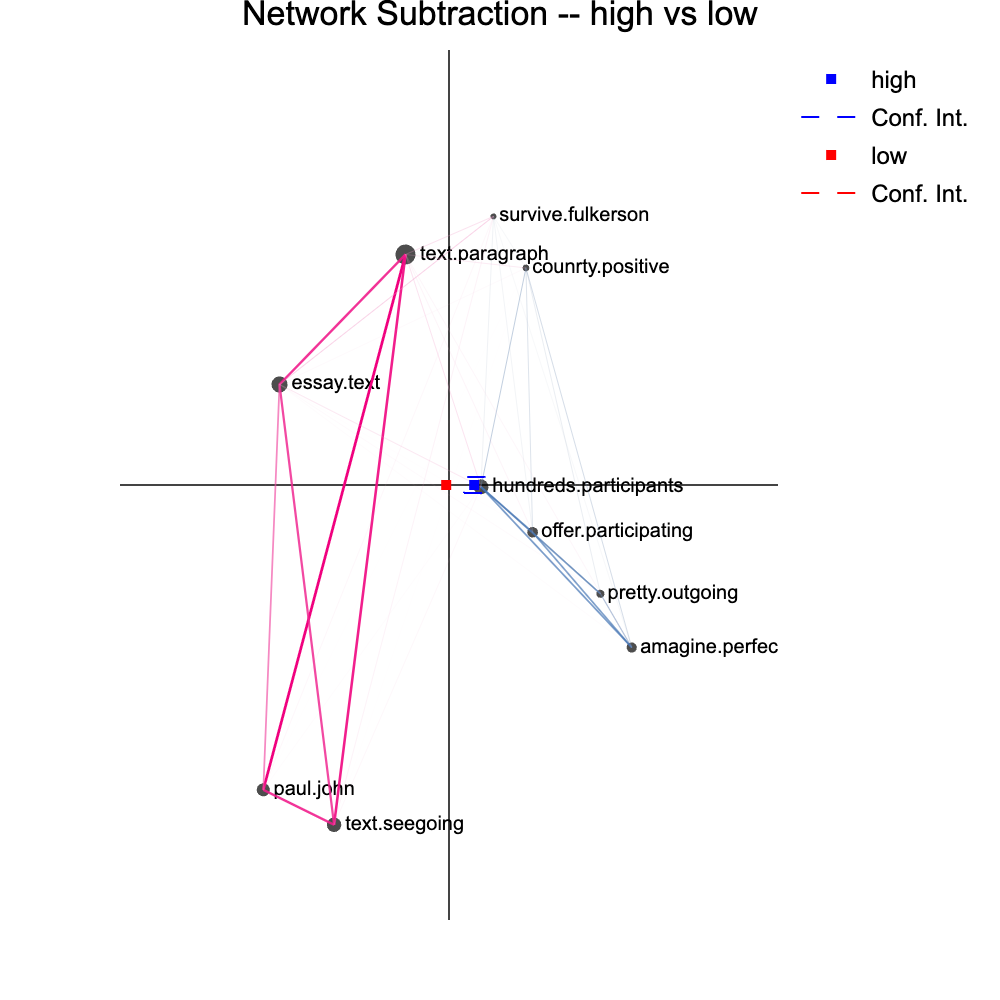}
        \caption{\textit{\textbf{Assignment 4}}}
        \label{fig:case2_a4_mid}
    \end{subfigure}\hfill
    \begin{subfigure}[t]{0.32\linewidth}
        \centering
        \includegraphics[width=\linewidth]{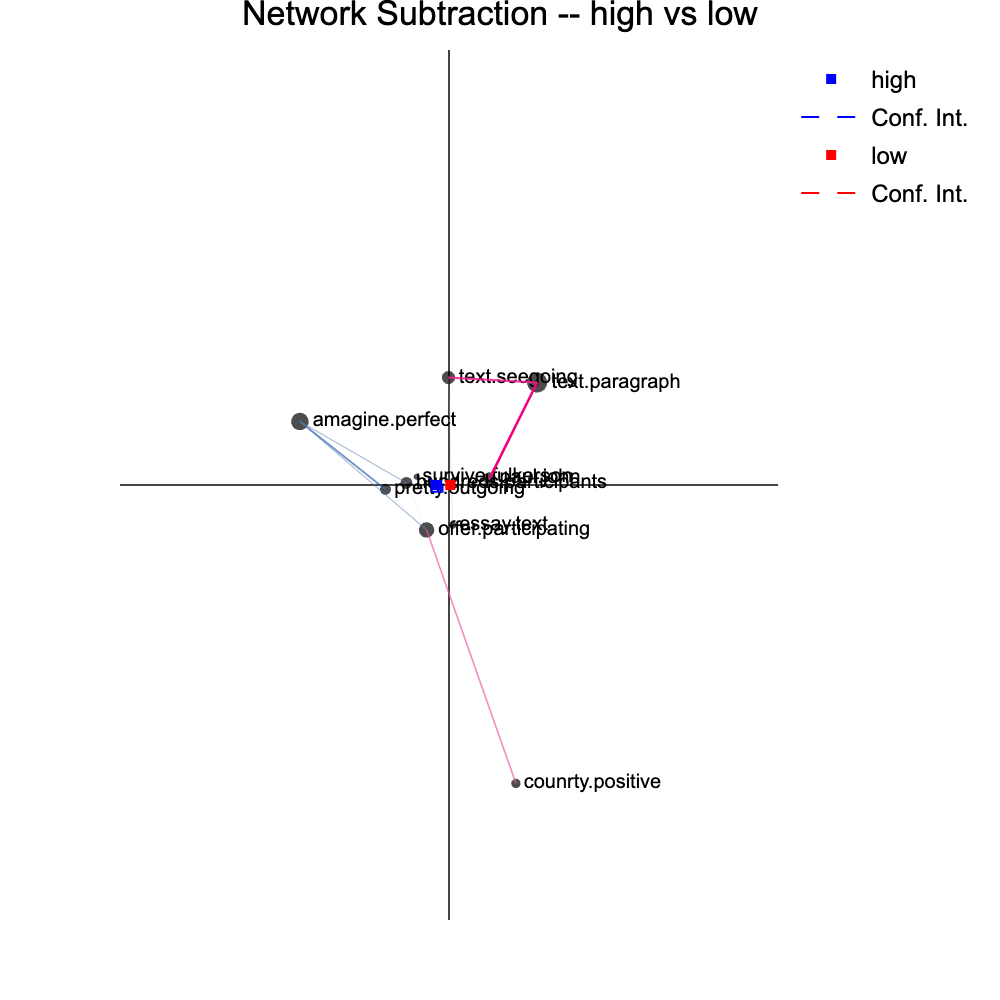}
        \caption{\textit{\textbf{Assignment 4}}}
        \label{fig:case2_a4_high}
    \end{subfigure}

    \medskip

    \begin{subfigure}[t]{0.32\linewidth}
        \centering
        \includegraphics[width=\linewidth]{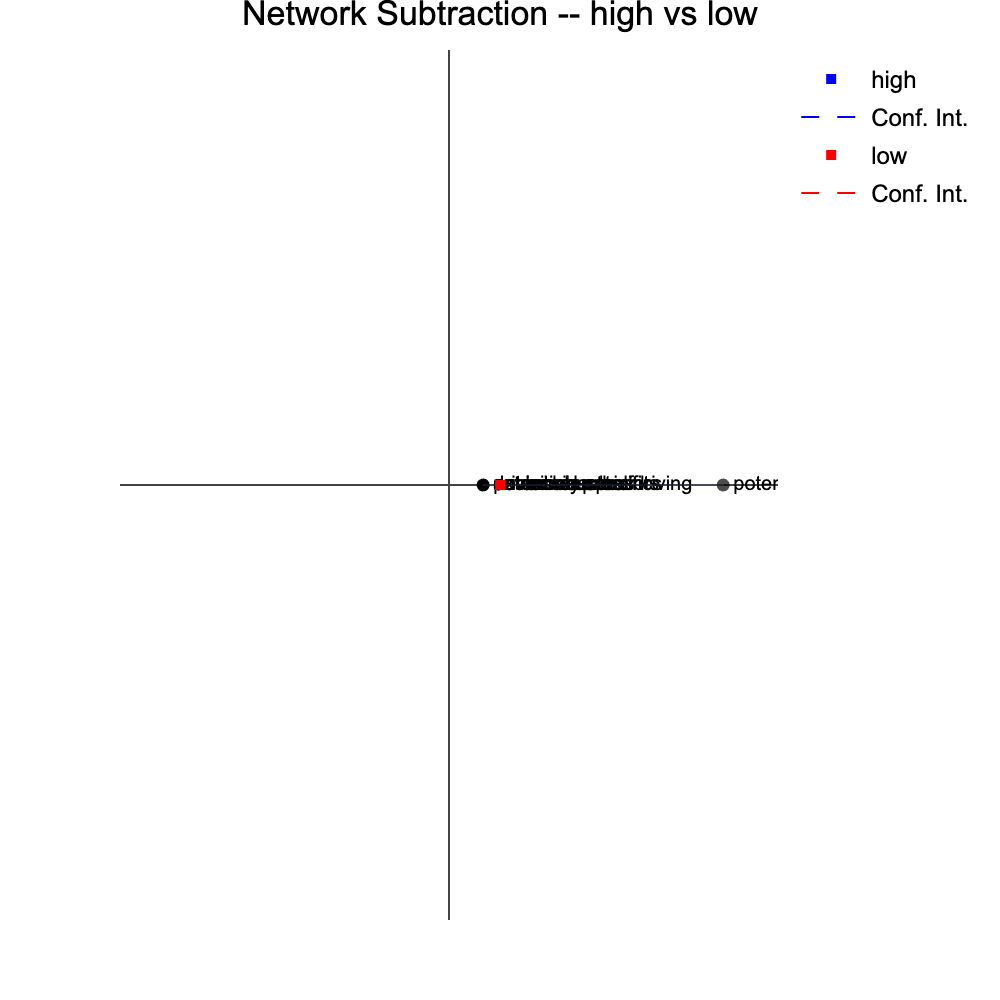}
        \caption{\textit{\textbf{Assignment 5}}}
        \label{fig:case2_a5_low}
    \end{subfigure}\hfill
    \begin{subfigure}[t]{0.32\linewidth}
        \centering
        \includegraphics[width=\linewidth]{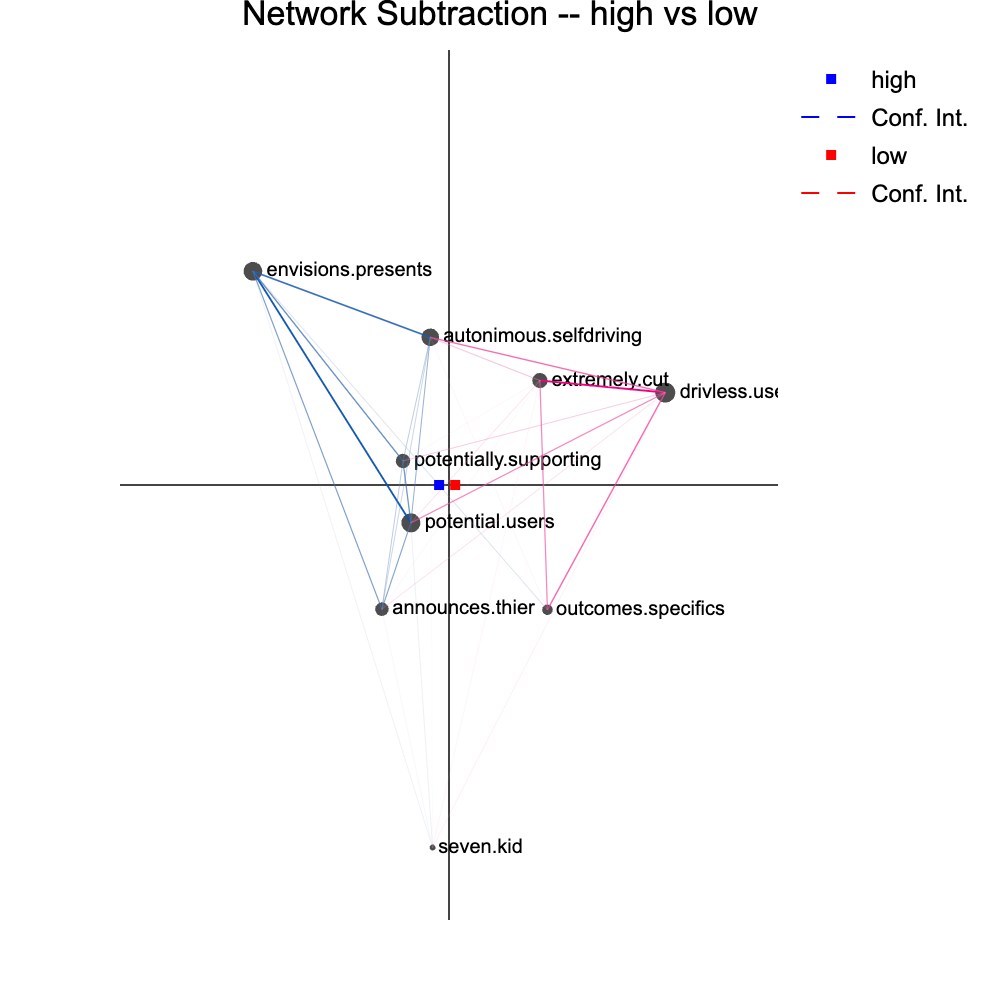}
        \caption{\textit{\textbf{Assignment 5}}}
        \label{fig:case2_a5_mid}
    \end{subfigure}\hfill
    \begin{subfigure}[t]{0.32\linewidth}
        \centering
        \includegraphics[width=\linewidth]{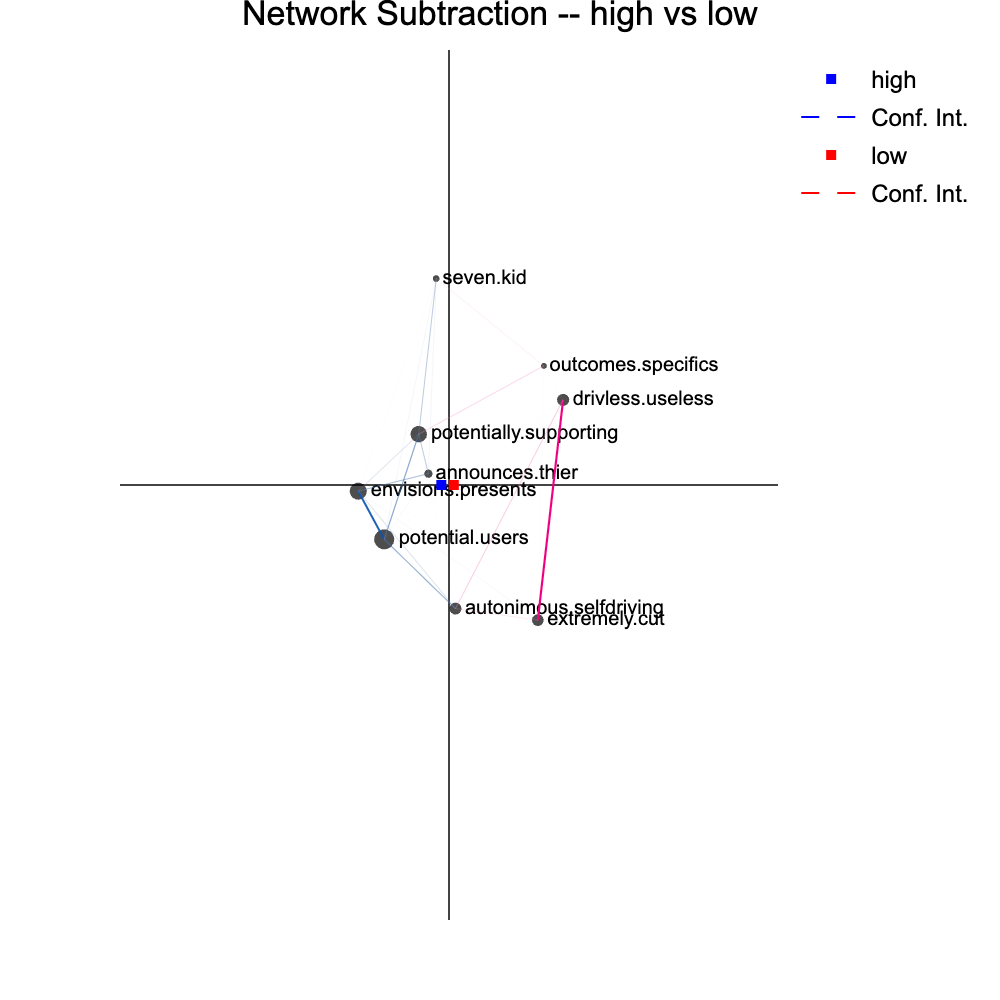}
        \caption{\textit{\textbf{Assignment 5}}}
        \label{fig:case2_a5_high}
    \end{subfigure}

    \caption{ENA results under different topic inclusion thresholds (low, medium, and high) in \textbf{Case 2}}
    \label{fig:case2_result_2x3}
\end{figure*}

As shown in Figures~\ref{fig:case2_a4_low}, \ref{fig:case2_a4_mid}, and \ref{fig:case2_a4_high}, \textit{\textbf{Assignment 4}}, which contains fewer utterances, is analyzed under low, medium, and high \texttt{topic\_inclusion\_th} settings, respectively. As the result, A low probability threshold caused most documents to be assigned to many topics simultaneously. This substantially increased co-occurrence strength between documents, resulting in very dense ENA networks. Under this condition, structural differences between the high-score and low-score groups were largely obscured. When a medium threshold was applied, each document was typically assigned to a limited number of topics. This produced a more balanced level of co-occurrence and allowed clearer structural differences between the two groups to emerge in the ENA difference networks. In contrast, a high threshold caused many documents to be assigned to very few topics or none at all. This led to sparse co-occurrence relations and reduced the interpretability of the ENA networks.

A similar pattern was observed in \textit{\textbf{Assignment 5}}, as shown in Figures~\ref{fig:case2_a5_low}, \ref{fig:case2_a5_mid}, and \ref{fig:case2_a5_high}. Low threshold settings produced overly dense networks that masked group differences, while high threshold settings resulted in sparse networks with limited structural information. Medium threshold settings achieved a better balance between multi-topic assignment and co-occurrence strength, allowing clear and interpretable differences between the high-score and low-score groups.

Overall, the results of \textbf{Case 2} indicate that the topic inclusion threshold plays a critical role in TopicENA analysis. Thresholds that are too low or too high both reduce the interpretability of ENA networks. Therefore, the selection of \texttt{topic\_inclusion\_th} should be guided by the distribution of word-level topic scores produced by BERTopic. Choosing a threshold from the middle range of the score distribution helps ensure reasonable multi-topic assignment and improves the clarity and validity of ENA results.

    


\subsection{Case 3: Scalability of TopicENA in Large-Scale Discourse Analysis}

In \textbf{Case 3}, we applied TopicENA to the ASAP 2.0 dataset to assess its scalability and practicality for large-scale text analysis. The data set includes 7 writing assignments totaling 24,728 student essays. After sentence-level segmentation, 457,002 utterances were obtained. Each essay was assigned to a high-performance group (scores 4-6) or a low-performance group (scores 1-3). The groupings were used for subsequent topic modeling and epistemic network analysis.

\begin{figure}[ht]
  \includegraphics[width=\columnwidth]{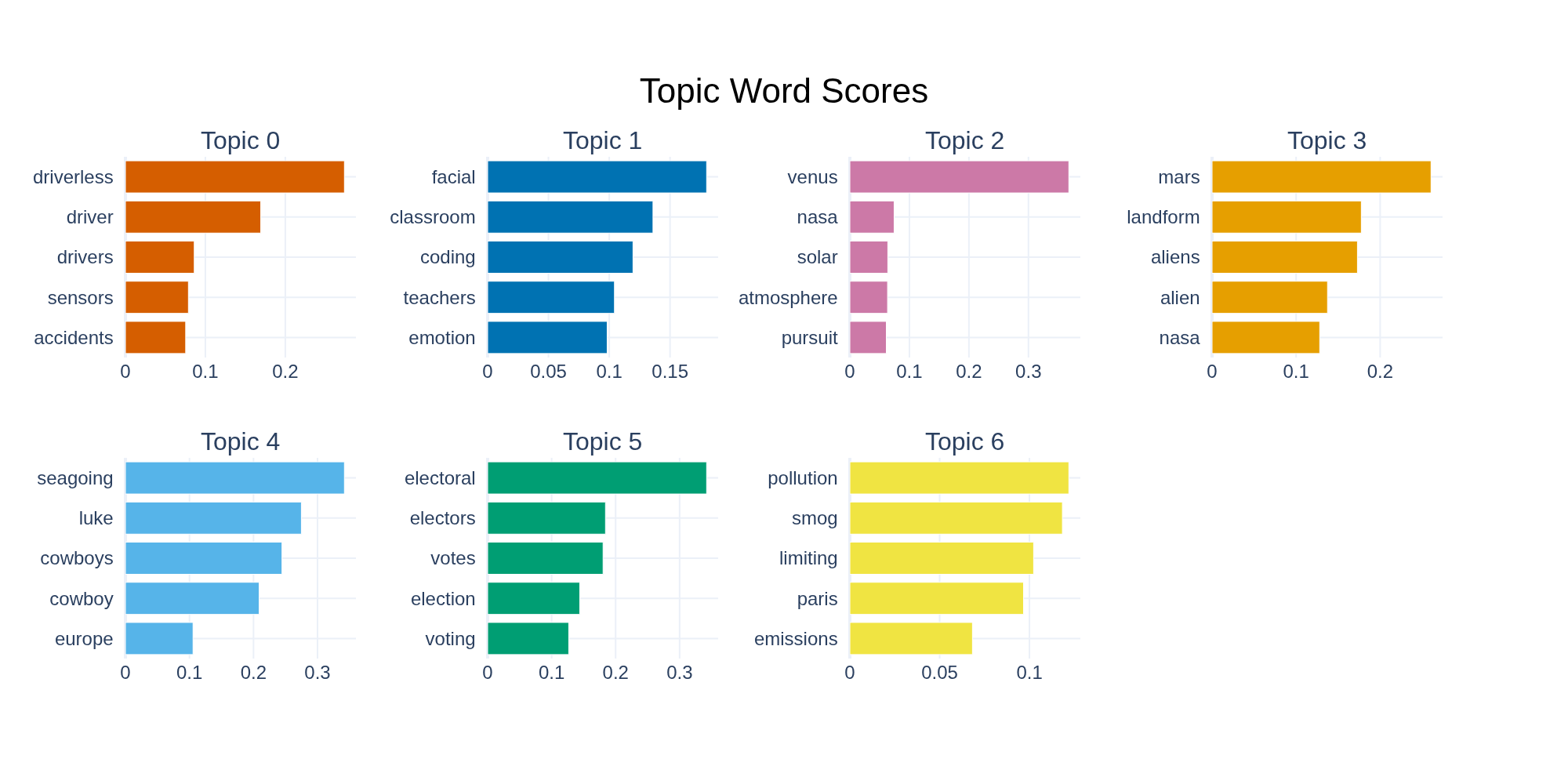}
  \caption{BERTopic results for \textbf{Case 3} showing seven topics identified from the full ASAP 2.0 dataset.}
  \label{fig:result_case3_topic}
\end{figure}

The results of BERTopic are shown in Figure~\ref{fig:result_case3_topic}. A total of seven topics were identified and labelled from topic 0 to topic 6. Each topic is represented by a set of high-weight keywords that form a clear and interpretable semantic pattern. Overall, the seven topics closely match the seven writing assignments in the ASAP 2.0 dataset, indicating that the semantic space of student essays is strongly determined by the writing tasks. Specifically, topic 0 is represented by keywords such as \texttt{driverless} and \texttt{driver}, which correspond to the assignment on \textit{Driverless Cars are Coming}. Topic 1 is characterized by \texttt{facial} and \texttt{classroom} and is in line with the assignment on the \textit{Making Mona Lisa Smile}. Topic 2 includes \texttt{venus} and \texttt{nasa} and corresponds to the assignment on the \textit{The Challenge of Exploring Venus}, while topic 3, defined by \texttt{mars} and \texttt{landform}, corresponds to the assignment that argues on the \textit{Unmasking the Face on Mars}. Topic 4 includes \texttt{seagoing} and \texttt{luke} and corresponds to the assignment about the \textit{A Cowboy Who Rode the Waves}.

Importantly, topic 5, characterized by \texttt{electoral} and \texttt{electors}, matches \textit{\textbf{Assignment \#4}}, whose writing prompt is provided in this Section~\ref{sec:experimental_procedure} and focuses on the Electoral College. Similarly, topic 6, defined by \texttt{pollution} and \texttt{smog}, aligns with \textit{\textbf{Assignment \#5}}, which addresses limiting car use to reduce pollution.

\begin{figure*}[t]
    \centering
    \begin{subfigure}[t]{0.33\linewidth}
        \centering
        \includegraphics[width=\linewidth]{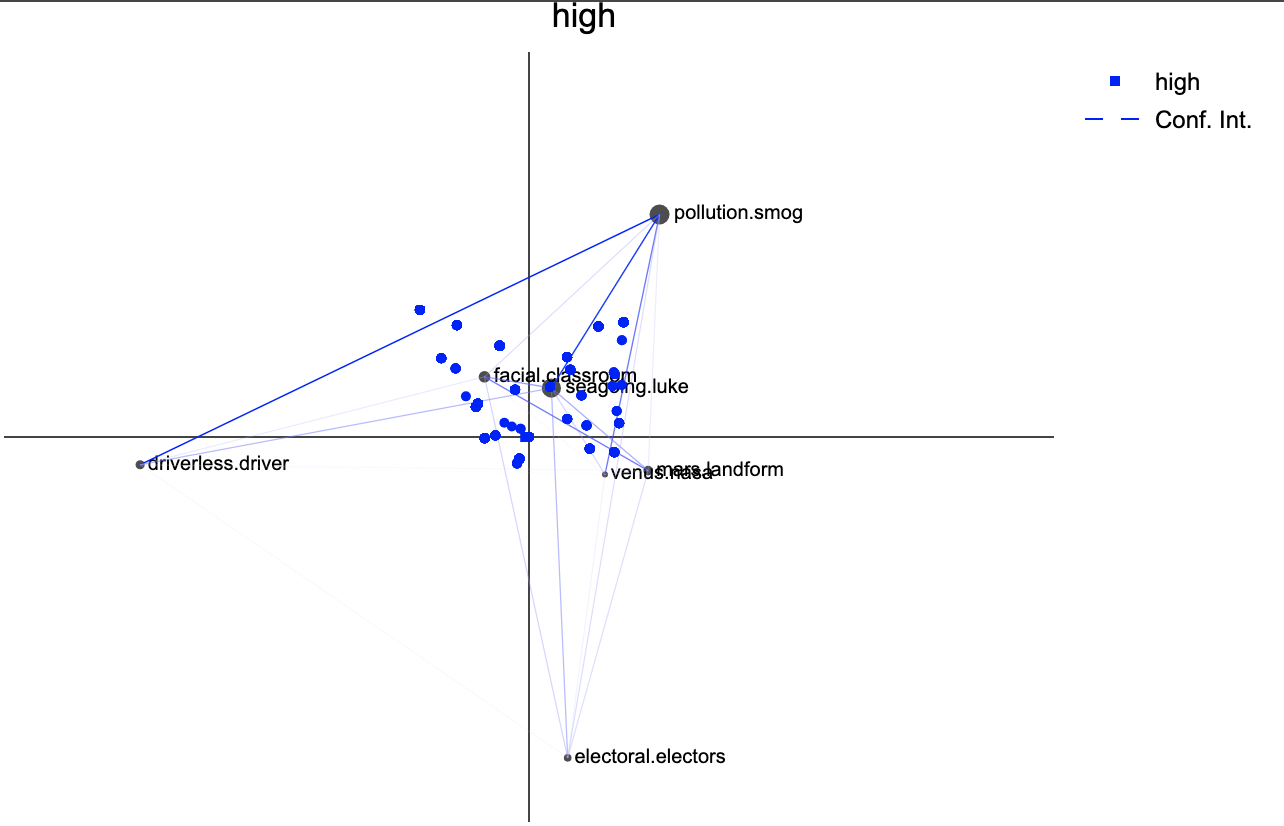}
        \caption{High performance group}
        \label{fig:case3_high}
    \end{subfigure}\hfill
    \begin{subfigure}[t]{0.33\linewidth}
        \centering
        \includegraphics[width=\linewidth]{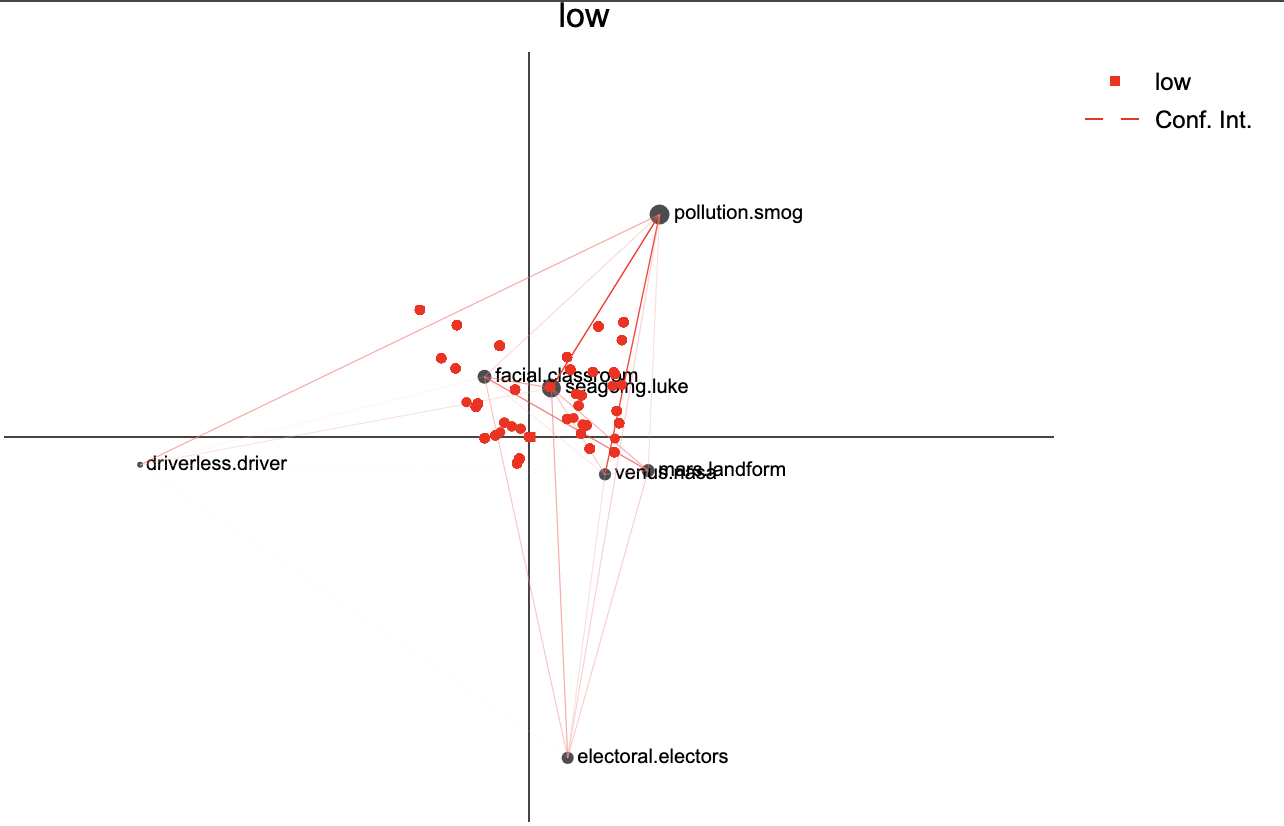}
        \caption{Low performance group}
        \label{fig:case3_low}
    \end{subfigure}\hfill
    \begin{subfigure}[t]{0.33\linewidth}
        \centering
        \includegraphics[width=\linewidth]{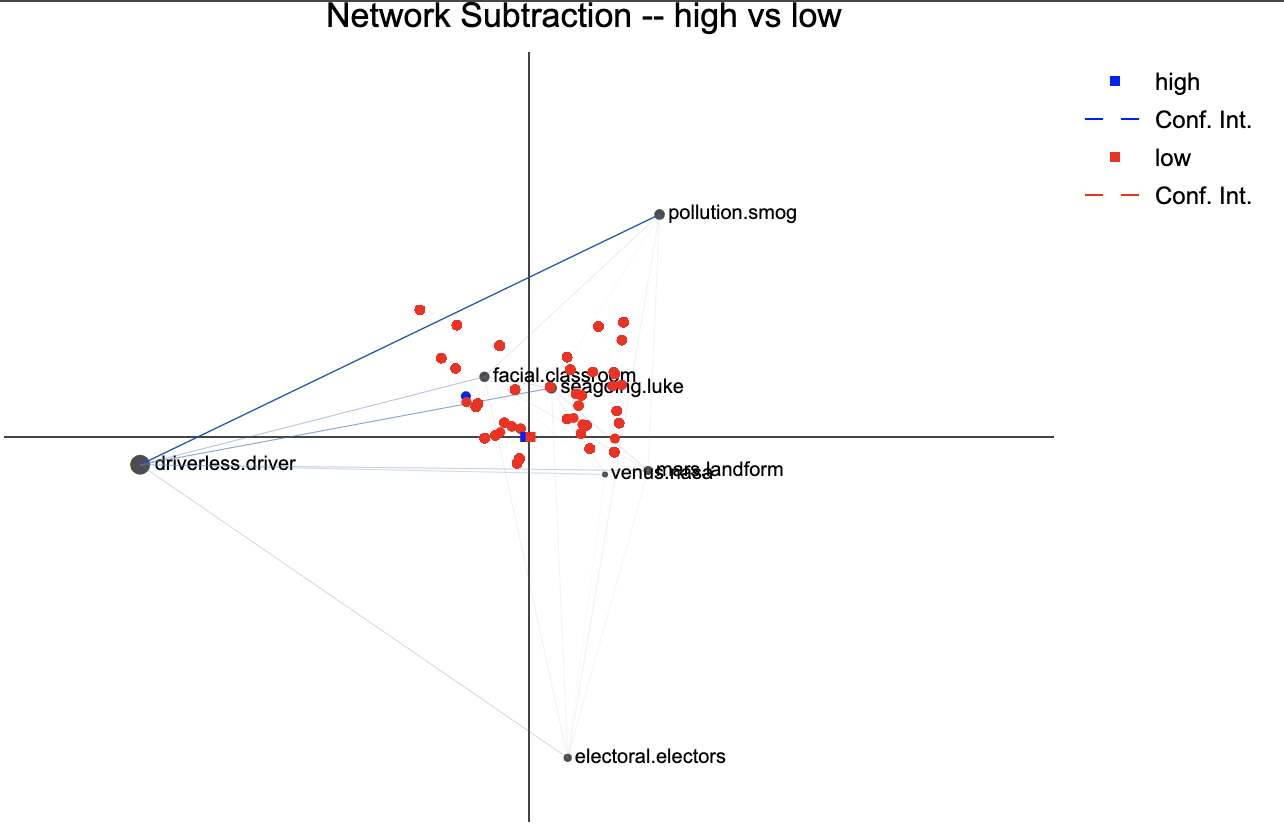}
        \caption{High–low subtraction}
        \label{fig:case3_sub}
    \end{subfigure}

    \caption{Epistemic networks for \textbf{Case 3}.}
    \label{fig:result_case3_ena}
\end{figure*}

The results of the epistemic network analysis for \textbf{Case 3} are shown in Figure~\ref{fig:result_case3_ena}. This figure summarizes the ENA outcomes from the full ASAP 2.0 dataset after topic-based semantic coding. The group-level epistemic networks for \textbf{Case 3} are shown in Figure~\ref{fig:case3_high} for the high-performance group and Figure~\ref{fig:case3_low} for the low-performance group. The high-performance network is displayed in blue, while the low-performance network is displayed in red. In both networks, nodes represent semantic concepts derived from the BERTopic results and correspond to the identified topics. Edges represent co-occurrence relationships between concepts within utterances. Node size indicates the overall strength of a concept’s connections with other concepts in the network, while edge thickness reflects the strength of co-occurrence between two connected concepts. The node labelled \texttt{driverless.driver} appears as a large node in the high-performance network but as a much smaller node in the low-performance network, showing a stronger overall connectivity for this concept in the high-performance group. In contrast, the node \texttt{electoral.electors} appear relatively small in the high-performance network but noticeably larger in the low-performance network. In addition, the node \texttt{pollution.smog} appears as a large node in both the high and low performance networks, signifying strong overall connectivity in both groups.

Second, Figure~\ref{fig:case3_sub} presents the subtract network, which highlights differences in connection strength between the high- and low-performing groups. In this network, only blue edges appear, showing stronger connections within the high-performance group. There are no red edges, meaning no connections are stronger in the low-performance group. The size of each node shows how much each concept’s connections differ between the two groups. The edge between \texttt{driverless.driver} and \texttt{pollution.smog} stands out as a prominent blue edge, indicating a stronger co-occurrence for this pair in the high-performance group.

To sum up, the results of \textbf{Case 3} show that TopicENA can identify semantic concepts that are closely consistent with the seven writing assignments in the ASAP 2.0 dataset, even without using assignment labels. This indicates that TopicENA can reliably capture task-related semantic patterns in large-scale, task-constrained writing data. Because the seven assignments are designed to be independent and each essay focuses on a single task, few cross-topic connections are observed in the overall epistemic networks. This pattern is also reflected in the difference network, where only a very limited number of connections appear. Notably, the only visible connection in the difference network links concepts related to driverless cars and environmental pollution, suggesting that high-performing essays tend to employ similar descriptive patterns when discussing these topics. Such a connection is not clearly observed in the low-performing group. Overall, the results show that TopicENA can both recover assignment-level semantic structure and highlight a small number of meaningful structural differences under strongly separated task conditions.



        

    
  

\section{Discussion}

The findings from \textbf{Case 1} and \textbf{Case 2} demonstrate that a meaningful and interpretable knowledge network does not emerge automatically; instead, it depends on researchers’ informed choices regarding the dataset. \textbf{Case 1} illustrates that topic granularity should be treated as a theoretical hypothesis rather than a purely technical parameter. Notably, applying the same granularity setting across different task contexts does not guarantee the emergence of a similarly clear network structure. For smaller datasets with a focused conceptual scope, finer granularity facilitates the distinction of concepts. In contrast, applying the same granularity to larger datasets and broader conceptual spaces can result in overly dense networks that may collapse. Thus, a first guiding principle for parameter adjustment is that dataset scale should be considered when selecting semantic granularity: larger datasets often benefit from relatively coarser semantic abstraction, whereas smaller datasets may support finer-grained representations. This approach ensures that the level of semantic abstraction aligns with both the characteristics of the task and the size of the data.

However, appropriate granularity settings alone are insufficient to ensure the formation of a coherent network. \textbf{Case 2} further demonstrates that the topic inclusion threshold critically influences which semantic units are incorporated into the construction of knowledge structures. Setting the threshold too low admits an excessive number of semantic units, resulting in a dense and less interpretable network. Conversely, a threshold set too high may exclude significant semantic signals, leading to the absence of discernible structure. In \textbf{Case 2}, a probability threshold of 0.05 indicated stable structural differences, but this value does not possess universal applicability. Instead, it suggests an effective range in which semantic units are neither overrepresented nor prematurely excluded. This leads to a second principle: the topic inclusion threshold should be determined based on actual topic probabilities or vocabulary score distributions, rather than being fixed in advance. In summary, the knowledge structure in TopicENA does not emerge naturally from the data; it becomes apparent only when semantic abstraction levels and inclusion criteria are jointly calibrated to align with the specific task and data characteristics. Especially in at-scale analysis, structural stability arises from the coordination of multiple analytical conditions rather than from any single optimal parameter setting.

This study advances the scale of ENA to an unprecedented level. For example, \citet{gavsevic2019sens}’s application of LDA topic modeling involved approximately 6,000 posts. In contrast, \textit{\textbf{Assignment 4}} in the ASAP 2.0 dataset, though considered smaller, contains over 37,000 utterances, while \textit{\textbf{Assignment 5}}, which includes a 1,959-word essay, comprises nearly 120,000 utterances. Such scale has not been previously reported in ENA studies. Manual coding would be impractical at this scale, underscoring the need for automated topic modeling. BERTopic offers several advantages over LDA. The primary distinction lies in the treatment of contextual semantics versus the bag-of-words approach. LDA models discourse based on word frequency, disregarding order and context, whereas BERTopic derives semantics from context, allowing the same concept to be expressed in varied linguistic forms. This capability enables BERTopic to maintain semantic consistency across both sentence-level and utterance-level analyses. In contrast, LDA may generate fractional or mixed topics. Furthermore, BERTopic topics are more closely aligned with conceptual categories than with mere vocabulary groupings. Finally, since ENA nodes represent "concepts," topics generated by LDA, which are essentially collections of high-frequency words, often require human interpretation to reach the conceptual level. BERTopic topics, constructed within a semantic space, are more readily mapped to interpretable semantic units.

TopicENA should be regarded not as a tool for the automatic generation of epistemic networks, but as a conceptual framework that redefines the researcher’s role. By facilitating large-scale, systematic exploration across various granularities and threshold settings, TopicENA shifts human involvement from labor-intensive, instance-level annotation to higher-order interpretation, comparison, and theoretical sense-making of emergent frameworks.

Another distinction from manual coding is that, in manually coded ENA, theoretical assumptions are embedded throughout the analytic process, including the development of the coding scheme, coder training, and the establishment of inter-coder agreement. Although the resulting epistemic structure may appear to emerge organically from the data, the coding framework has, in fact, already constrained and predetermined the possible forms that such structures can take.

In contrast, TopicENA does not embed theory at the level of individual annotations. Instead, theoretical assumptions are articulated through analytic configurations, such as the selection of semantic granularity and inclusion thresholds. These assumptions do not directly dictate the resulting structure; rather, they establish the conditions under which structure may or may not emerge. Additionally, these assumptions are enumerable, repeatable, and falsifiable. The resulting structure is not determined in a single step, but must persist across varying analytical conditions. This approach renders TopicENA a more methodologically transparent analytical framework.

\section{Limitation}

This study applied automated topic modeling combined with ENA without relying on manual coding as a reference.  Therefore, code-level reliability and validity checks were not applicable. This was an intentional design choice rather than a methodological weakness. As noted by~\citet{chen2023we}, topic modeling is not intended to replicate human expert coding, but to support scalable semantic exploration of large text corpora. Accordingly, TopicENA does not aim to mimic human judgments at the level of individual text units, but to use automatically generated semantic units to examine learners’ knowledge structures and patterns of concept co-occurrence. In this context, the role of human experts shifts from low-level coding agreement to higher-level conceptual and structural interpretation. Nevertheless, the semantic units used in TopicENA are data-driven rather than predefined theoretical constructs, and using thematic probabilities to construct co-occurrence might be affected by model parameters and the corpus's features, in turn affecting the stability and interpretability of the established network.

\section{Conclusion}

This study proposed TopicENA, an analytical tool that combines TopicBERT and ENA, aiming to move beyond the heavy reliance on manual coding of ENA and enable its application to large-scale textual data. TopicENA use automated generated semantic themes to replace manual work, and shifts the role of researchers from instance-level text annotation to higher-level structural interpretation and comparison. This study has proven that using ENA to conducted structual analysis can be without manual coding. In actual practice, the thought topic probability threshold, TopicENA allow single text connect to several topics, and builds the topic co-occurrence structure for ENA analysis, which conserves the multi-aspect semantic features of discourses and avoids the constrained structural representation due to the one-to-one coding scheme, where each text segment is linked to a single concept. 

The experiment has shown the availability of data at different scales. In small, medium, and big data (Cases 1-3), can produce explainable epistemic networks and consistent semantic structural tendencies. Which means TopicENA has a certain extent of scalability and feasibility, and has extended the application potential of ENA in large-scale discourse analysis, learning analysis, and AI-assisted educational research. 

The study does not use manual coding as reference, for TopicENA is not for simulating human judgment, but used for exploratory semantic structure, support large scale analysis and group comparison. In future work, the impact of the topic parameter on network stability should be further investigated. Also, explore the potential complementarity with theory-driven coding and its application in a multilingual scenario.

\bibliography{custom}

@article{janssens2025comparative,
  title={A Comparative Analysis of Topic Reduction Techniques for BERTopic},
  author={Janssens, Wannes and Bogaert, Matthias and Van den Poel, Dirk},
  journal={IEEE Access},
  volume={13},
  pages={204087--204103},
  year={2025},
  publisher={IEEE}
}

@article{raman2024unveiling,
  title={Unveiling the dynamics of AI applications: A review of reviews using scientometrics and BERTopic modeling},
  author={Raman, Raghu and Pattnaik, Debidutta and Hughes, Laurie and Nedungadi, Prema},
  journal={Journal of Innovation \& Knowledge},
  volume={9},
  number={3},
  pages={100517},
  year={2024},
  publisher={Elsevier}
}

@article{kherwa2020topic,
  title={Topic modeling: a comprehensive review.},
  author={Kherwa, Pooja and Bansal, Poonam},
  journal={EAI Endorsed Trans. Scalable Inf. Syst.},
  volume={7},
  number={24},
  pages={e2},
  year={2020}
}

@article{chen2023we,
  title={What we can do and cannot do with topic modeling: A systematic review},
  author={Chen, Yingying and Peng, Zhao and Kim, Sei-Hill and Choi, Chang Won},
  journal={Communication Methods and Measures},
  volume={17},
  number={2},
  pages={111--130},
  year={2023},
  publisher={Taylor \& Francis}
}

@article{grimmer2013text,
  title={Text as data: The promise and pitfalls of automatic content analysis methods for political texts},
  author={Grimmer, Justin and Stewart, Brandon M},
  journal={Political analysis},
  volume={21},
  number={3},
  pages={267--297},
  year={2013},
  publisher={Cambridge University Press}
}

@article{blei2012probabilistic,
  title={Probabilistic topic models},
  author={Blei, David M},
  journal={Communications of the ACM},
  volume={55},
  number={4},
  pages={77--84},
  year={2012},
  publisher={ACM New York, NY, USA}
}

@article{blei2003latent,
  title={Latent dirichlet allocation},
  author={Blei, David M and Ng, Andrew Y and Jordan, Michael I},
  journal={Journal of machine Learning research},
  volume={3},
  number={Jan},
  pages={993--1022},
  year={2003}
}

@article{pantic2022making,
  title={Making sense of teacher agency for change with social and epistemic network analysis},
  author={Panti{\'c}, Nata{\v{s}}a and Galey, Sarah and Florian, Lani and Joksimovi{\'c}, Sre{\'c}ko and Viry, Gil and Ga{\v{s}}evi{\'c}, Dragan and Knutes Nyqvist, Hel{\'e}n and Kyritsi, Krystallia},
  journal={Journal of educational change},
  volume={23},
  number={2},
  pages={145--177},
  year={2022},
  publisher={Springer}
}

@inproceedings{vega2021negotiating,
  title={Negotiating tensions: A study of pre-service English as foreign language teachers’ sense of identity within their community of practice},
  author={Vega, Hazel and Irgens, Golnaz Arastoopour and Bailey, Cinamon},
  booktitle={International Conference on Quantitative Ethnography},
  pages={277--291},
  year={2021},
  organization={Springer}
}

@inproceedings{ko2024exploring,
  title={Exploring students’ changing conceptions about eLearning leadership},
  author={Ko, Pakon and Law, Nancy and Liu, Cong},
  booktitle={International Conference on Quantitative Ethnography},
  pages={202--216},
  year={2024},
  organization={Springer}
}

@inproceedings{paquette2021using,
  title={Using epistemic networks to analyze self-regulated learning in an open-ended problem-solving environment},
  author={Paquette, Luc and Grant, Theodore and Zhang, Yingbin and Biswas, Gautam and Baker, Ryan},
  booktitle={International conference on quantitative ethnography},
  pages={185--201},
  year={2021},
  organization={Springer}
}

@article{shaffer2016tutorial,
  title={A tutorial on epistemic network analysis: Analyzing the structure of connections in cognitive, social, and interaction data},
  author={Shaffer, David Williamson and Collier, Wesley and Ruis, Andrew R},
  journal={Journal of learning analytics},
  volume={3},
  number={3},
  pages={9--45},
  year={2016}
}

@article{tu2025effects,
  title={Effects on the learning achievement, approaches to learning, and multi-stage reflection quality of students with different levels of digital self-efficacy in a data literacy course: An ARCS-based self-reflective online learning model},
  author={Tu, Yun-Fang and Hwang, Gwo-Jen and Hu, Dongpin},
  journal={Computers \& Education},
  volume={238},
  pages={105397},
  year={2025},
  publisher={Elsevier}
}

@article{chang2025generative,
  title={Generative AI-assisted reflective writing for improving students’ higher order thinking},
  author={Chang, Ching-Yi and Lin, Hui-Chen and Yin, Chengjiu and Yang, Kai-Hsiang},
  journal={Educational Technology \& Society},
  volume={28},
  number={1},
  pages={270--285},
  year={2025},
  publisher={JSTOR}
}

@article{crossley2025large,
  title={A large-scale corpus for assessing source-based writing quality: ASAP 2.0},
  author={Crossley, Scott A and Baffour, Perpetual and Burleigh, L and King, Jules},
  journal={Assessing Writing},
  volume={65},
  pages={100954},
  year={2025},
  publisher={Elsevier}
}

@misc{learning-agency-lab-automated-essay-scoring-2,
    author = {Scott Crossley and Perpetual Baffour and Jules King and Lauryn Burleigh and Walter Reade and Maggie Demkin},
    title = {Learning Agency Lab - Automated Essay Scoring 2.0},
    year = {2024},
    howpublished = {\url{https://kaggle.com/competitions/learning-agency-lab-automated-essay-scoring-2}},
    note = {Kaggle}
}

@article{gavsevic2019sens,
  title={SENS: Network analytics to combine social and cognitive perspectives of collaborative learning},
  author={Ga{\v{s}}evi{\'c}, Dragan and Joksimovi{\'c}, Sre{\'c}ko and Eagan, Brendan R and Shaffer, David Williamson},
  journal={Computers in Human Behavior},
  volume={92},
  pages={562--577},
  year={2019},
  publisher={Elsevier}
}

@inproceedings{cai2017epistemic,
  title={Epistemic network analysis and topic modeling for chat data from collaborative learning environment},
  author={Cai, Zhiqiang and Eagan, Brendan and Dowell, Nia and Pennebaker, J and Shaffer, D and Graesser, A},
  booktitle={Proceedings of the 10th international conference on educational data mining},
  year={2017}
}

@article{grootendorst2022bertopic,
  title={BERTopic: Neural topic modeling with a class-based TF-IDF procedure},
  author={Grootendorst, Maarten},
  journal={arXiv preprint arXiv:2203.05794},
  year={2022}
}

\appendix



\end{document}